\pdfoutput=1

\documentclass[11pt]{article}

\usepackage[]{acl}

\usepackage{times}
\usepackage{latexsym}

\usepackage[T1]{fontenc}

\usepackage[utf8]{inputenc}

\usepackage{microtype}

\usepackage{inconsolata}

\usepackage{graphicx}

\newcommand*{\eg}{{\em e.g.}}

\newcommand*{\ie}{{\em i.e.}}

\newcommand{\ap}[1]{\textbf{\textcolor{blue}{#1}}}
\usepackage{graphicx}
\usepackage{amsmath}
\usepackage{soul}
\usepackage{pgf-pie}
\usepackage{tabularx}
\usepackage[T1]{fontenc}
\usepackage[tableposition=top]{subcaption}
\usepackage{multirow}
\usepackage{multicol}
\usepackage{booktabs}
\usepackage{tikz}
\usepackage{pgfplots}

\usepgfplotslibrary{colorbrewer}
\pgfplotsset{compat = 1.15, cycle list/Set1-8} 
\usetikzlibrary{pgfplots.statistics, pgfplots.colorbrewer} 
\usepackage{pgfplotstable}
\usepackage{filecontents}
\usepackage{varwidth}
\usepackage{soul}
\usepackage[export]{adjustbox}
\usepackage{dblfloatfix}

\usetikzlibrary{pgfplots.groupplots}
\usetikzlibrary{shapes.geometric}
\usetikzlibrary{positioning,fit,shapes.geometric,backgrounds, calc}
\usetikzlibrary{arrows,decorations.markings}
\usetikzlibrary{shapes.arrows}
\usetikzlibrary{patterns}
\tikzset{textnode/.style={inner sep=0pt,outer sep=0,execute at begin node={\strut}}}
\tikzstyle{state} = [textnode,circle, draw, inner sep=0pt, outer sep=0]
\usepgfplotslibrary{groupplots}

\pgfplotsset{every axis/.append style={
                    xlabel={$x$},          
                    ylabel={$y$},          
                    label style={font=\sffamily},
                    tick label style={font=\sffamily\scriptsize},
                    xticklabel style = {font=\sffamily\scriptsize},
                    title style = {font=\small\sffamily},
                    ylabel near ticks,
                    y label style={font=\sffamily\scriptsize},
                    xlabel near ticks,
                    x label style={font=\sffamily\scriptsize},
                    legend cell align={left},
                    legend style={draw=none, font=\sffamily\scriptsize},
                    },
                    legend image code/.code={
                    \draw[mark repeat=2,mark phase=2]
                        plot coordinates {
                        (0cm,0cm)
                        (0.15cm,0cm)        
                        (0.3cm,0cm)         
                        };%
                    }
                    }
\pgfplotsset{compat=newest}

\makeatletter
\pgfplotsset{
    boxplot prepared from table/.code={
        \def\tikz@plot@handler{\pgfplotsplothandlerboxplotprepared}%
        \pgfplotsset{
            /pgfplots/boxplot prepared from table/.cd,
            #1,
        }
    },
    /pgfplots/boxplot prepared from table/.cd,
        table/.code={\pgfplotstablecopy{#1}\to\boxplot@datatable},
        row/.initial=0,
        make style readable from table/.style={
            #1/.code={
                \pgfplotstablegetelem{\pgfkeysvalueof{/pgfplots/boxplot prepared from table/row}}{##1}\of\boxplot@datatable
                \pgfplotsset{boxplot/#1/.expand once={\pgfplotsretval}}
            }
        },
        make style readable from table=lower whisker,
        make style readable from table=upper whisker,
        make style readable from table=lower quartile,
        make style readable from table=upper quartile,
        make style readable from table=median,
        make style readable from table=lower notch,
        make style readable from table=upper notch,
                make style readable from table=average
}
\makeatother

\newcommand{\computerowindex}[2]{
\def\xindex{-1}
\pgfplotstableforeachcolumnelement{x}\of#1\as\cell{%
\edef\cellname{#2}
    \ifx\cell\cellname\let\xindex\pgfplotstablerow\fi   
}
}
\newcommand{\boxplotprepared}[2]{
\computerowindex{#1}{#2}
\expandafter\boxplotpreparedNum\expandafter#1\expandafter{\xindex}
}

\newcommand{\boxplotpreparedNum}[2]{
\addplot[boxplot prepared from table={
    table=#1,
    row=#2,
    average=avg,
    lower whisker = firstmin,
    upper whisker = firstmax,
    lower quartile = q1,
    upper quartile = q3,
    median=q2,
    },
    boxplot prepared]
coordinates {};
}

\title{Digital Gatekeepers: Google’s Role in Curating Hashtags and Subreddits}

\author{
    Amrit Poudel\textsuperscript{\rm 1}
    Yifan Ding\textsuperscript{\rm 1},
    Jürgen Pfeffer\textsuperscript{\rm 2},
    Tim Weninger\textsuperscript{\rm 1} \\
    \textsuperscript{\rm 1}Department of Computer Science and Engineering, University of Notre Dame, Notre Dame, USA\\
    \textsuperscript{\rm 2}School of Social Science and Technology, Technical University of Munich, Munich, Germany\\
    \{apoudel, yding, tweninge\}@nd.edu,  
    \{juergen.pfeffer\}@tum.de \\
}

\begin{document}
\maketitle
\begin{abstract}

Search engines play a crucial role as digital gatekeepers, shaping the visibility of Web and social media content through algorithmic curation. This study investigates how search engines like Google selectively promotes or suppresses certain hashtags and subreddits, impacting the information users encounter. By comparing search engine results with nonsampled data from Reddit and Twitter/X, we reveal systematic biases in content visibility. Google’s algorithms tend to suppress subreddits and hashtags related to sexually explicit material, conspiracy theories, advertisements, and cryptocurrencies, while promoting content associated with higher engagement. These findings suggest that Google’s gatekeeping practices influence public discourse by curating the social media narratives available to users.
\end{abstract}



\section{Introduction}

\begin{figure}[t]
    \centering
    \includegraphics[width=0.95\linewidth]{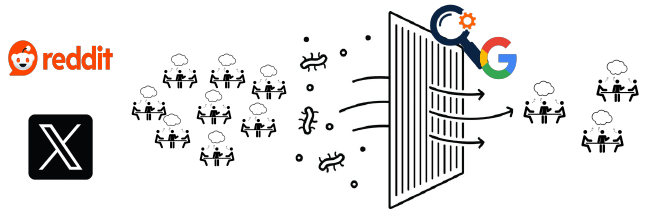}
    \caption{Search Engines curate and filter social media content before displaying results.}
    \label{fig:teaser}
\end{figure}

Online social media platforms, despite their limitations~\cite{ivan2015social} and potential risks~\cite{bert2016risks, abolfathi2022identification}, have revolutionized how individuals connect and communicate with others who share similar interests. The rapid growth in their usage can be attributed to the ubiquity of smartphones and advancements in social psychology and artificial intelligence~\cite{grandinetti2021examining}, which have transformed social media into a key driver of both individual interaction and public discourse. As the volume of social media content has surged, search engines have emerged as critical gatekeepers, filtering and mediating access to content from platforms like Reddit and Twitter/X~\cite{freelon2018computational}. However, this gatekeeping introduces potential biases that shape the visibility of subreddits and hashtags, influencing the flow of information and impacting public conversations as illustrated in Fig.~\ref{fig:teaser}. Research shows that biased search rankings can significantly affect consumer and voter decisions; one study found that such biases could shift voting preferences by over 20\% among undecided voters in the U.S. and India~\cite{epstein2015search}. This phenomenon, known as the \textit{search engine manipulation effect}, raises concerns about the role of dominant search engines in shaping democratic processes and underscores the importance of understanding how they curate online content.

To explore the framing effects of search engines, access to data is essential. However, the discontinuation of API access to social media sites have created significant barriers to obtaining this data. This period of data inaccessibility has been termed the \textit{Post-API era}~\cite{freelon2018computational, poudel2024navigating}, which has notably hindered research across various fields, including discourse analysis~\cite{de2014mental, stine2020comparative}, computational social science~\cite{priya2019should,hassan2020towards}, computational linguistics~\cite{basile2021dramatic,wang2021predicting,melton2021public,liu2020sentiment}, and human behavior studies~\cite{choi2015characterizing, thukral2018analyzing}, among others~\cite{weng2011event, sakaki2010earthquake}.

\begin{figure*}
    \centering
    \includegraphics[width=\textwidth]{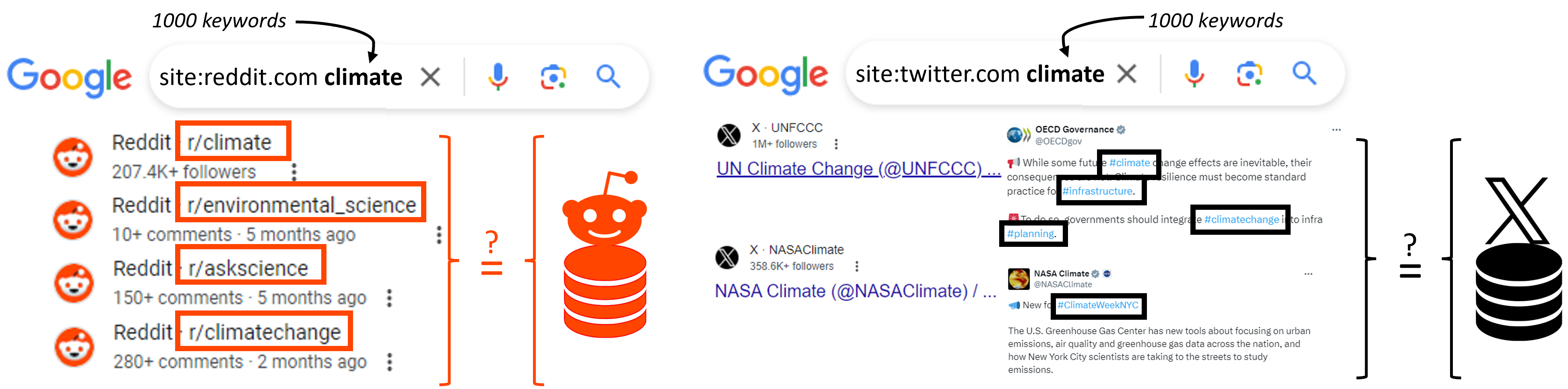}
    \caption{Comparison of Google search engine results with Reddit subreddits (left) and Twitter/X hashtags (right). The figure highlights differences in visibility and ranking across 1,000 random queries compared to nonsampled platform data, illustrating how search engines act as gatekeepers, influencing the prominence of online communities through filtering and moderation practices.}
    \label{fig:method}
\end{figure*}

\paragraph{Search Engine Result Pages.} Search engines frequently establish data-sharing agreements with social media platforms, allowing them access to large-scale, up-to-date data without the need for Web scraping. For instance, data from Google Trends can be used to calibrate and track the popularity of topics over time~\cite{west2020calibration}. In the \textit{Post-API era}, Search Engine Results Pages (SERPs) have emerged as a possible alternative data source for computing and social science research~\cite{scheitle2011google, young2018using, yang2015forecasting, pan2012forecasting}. However, as SERPs present results as paginated lists ranked by relevance, they inherently impose a layer of algorithmic moderation. This ranking process is central to the usability of search engines but also introduces biases in how content is prioritized, raising questions about the gatekeeping power of these platforms~\cite{sundin2022whose}.

\paragraph{Subreddits and Hashtags} 

Subreddits and hashtags are two examples of ways that platforms provide spaces for users with similar interests to gather and can even lead to the formation of new groups~\cite{krohn2022subreddit}. Other platforms like Facebook, WhatsApp, Telegram, and Weibo also support topical discussion or community formation in similar ways. 

Analysis of these dynamics has led to deep insights and countless studies on engagement, membership, conflict, and discourse both within specific groups and in general (\eg,~\cite{soliman2019characterization, weld2022makes, long2023just}). Continued study of these dynamics is predicated on the ability to gather data from these social platforms. In light of the new restrictions on social media data collection as well as the previous findings on bias in SERP data, the following questions arise: 

Our research builds upon previous work that investigates the \textit{page-level dynamics} of how individual posts or pages containing certain keywords are promoted or suppressed within search engine result pages (SERPs)~\cite{poudel2024navigating}. However, we take a broader \textit{community-based approach} that underscores the crucial role of subreddits and hashtags in shaping narratives. This shift in perspective allows us to uncover dimensions that are often overlooked in more granular studies. While we concur with prior research regarding the existence of bias in SERP representation, our findings extend this understanding by revealing how these biases operate at the community and topic levels. Search engine algorithms, we demonstrate, not only propagate bias but also significantly frame the larger narratives that emerge from online communities.

Building on these contributions, we turn our focus to the key research questions that guide our investigation. These questions aim to deepen our understanding of how search engines function as gatekeepers, shaping the visibility and framing of entire communities and the narratives they promote. By examining both the systemic biases that influence which subreddits and hashtags are surfaced or suppressed, and the broader implications of these dynamics for online discourse, the following three research questions seek to uncover the mechanisms through which search engines mediate public conversations.

\begin{enumerate}
\item How do search engine rankings and moderation policies serve as gatekeeping mechanisms that shape the visibility of subreddits and hashtags within online discourse?

\item How does the toxicity of content differ between subreddits and hashtags that appear in search engine result pages (SERP) and those that do not?

\item Which subreddits and hashtags are systematically promoted or suppressed by search engine algorithms and moderation practices, and what common characteristics can be identified among these topics and communities?

\end{enumerate}

To address these questions, and as illustrated in Fig~\ref{fig:method}, we compared the prevalence of subreddits and hashtags from non-sampled data obtained directly from Reddit and Twitter/X with those identified in thousands of SERPs from Google’s web search engine\footnote{We utilized the ScaleSERP service (\url{http://scaleserp.com})} during the same time period.

Overall, we find that Google significantly and dramatically biases the subreddits and hashtags that are returned in important (but not malicious or nefarious) ways. On Reddit, the subreddits that were most suppressed included r/AskReddit, r/AutoNewspaper, and r/dirtykikpals; on Twitter/X the hashtags that were most suppressed were \#voguegala2022xmileapo, \#nft, and \#nsfwtwt. Looking at the results broadly, we find that subreddits and hashtags that contain sexually explicit content, that promote conspiracy theories, that contain many advertisements, and that promote cryptocurrencies are less likely to be returned by Google compared to nonsampled social media data. On the other hand, we find that gaming and entertainment subreddits and hashtags are more likely to be returned by Google compared to nonsampled social media data.

\section{Related Work}

Here we review key literature on (1) the influential role of search engines in shaping public discourse, and (2) challenges in data collection in social media research. Investigating the framing role of search engines in shaping public discourse requires access to robust data. However, the process of data collection presents its own set of challenges.

\subsection{Search Engines as Gatekeepers}

Search engines play a pivotal role in shaping social discourse and curating information, fundamentally influencing public perceptions and narratives~\cite{makhortykh2021hey, introna2000shaping, epstein2015search, pan2007google}. This curation is not merely a passive reflection of user interest but an active process that can amplify certain viewpoints while marginalizing others~\cite{gerhart2004web, epstein2015search}. Researchers have noted that algorithms governing search engines and social media platforms function as gatekeepers, determining which content is visible and how it is shown~\cite{goldman2005search}. This is particularly important given the sheer volume of information available online, where users rely on search engines to navigate and filter relevant content from the noise.

The mechanics of gatekeeping within search engines involve both the selection and filtering of information based on various criteria, including relevance, popularity, and alignment with the users' prior behavior~\cite{brin1998anatomy, baeza1999modern, hannak2013measuring}. As they do their work, they can inadvertently reinforce societal biases and echo chambers, shaping users' understanding of issues in ways that reflect hidden biases rather than a neutral presentation of information~\cite{gillespie2020content,gillespie2010politics}. 

The implications of these algorithmic choices extend beyond individual users to impact the broader social dynamics. As platforms prioritize content that generates higher engagement, they risk skewing the discourse towards more sensational or polarizing material, which can further entrench echo chambers and reduce exposure to a broad range of perspectives~\cite{barbera2020social}.

In summary, as curators of information, search engines significantly affect how social issues are framed and discussed in modern public discourse. Their role as gatekeepers not only determines what information is accessible but also influences the narratives that emerge within society, making it a critical path for investigation.

\subsection{Data Collection Strategies}
The rise of social media has transformed the study of online behavior~\cite{myslin2013using, young2009extrapolating}, but recent restrictions on data access have forced researchers to explore alternative methods. These methods include data recalibration strategies, alternative data sharing mechanisms, and new data acquisition techniques. Social media data often suffers from sampling bias, such as Twitter’s \textit{garden-hose} versus \textit{fire-hose} feed~\cite{morstatter2013sample}. Researchers have developed methods to address this through data cleaning and recalibration, which correct noisy labels and adjust for incomplete data~\cite{ilyas2019data, west2020calibration, ford2023competition}.

With data collection services becoming more restricted, alternatives like data donation have emerged, where users voluntarily provide their data~\cite{carriere2023best, ohme2023digital}. Others propose policy-driven solutions, such as requiring platforms to share public data under regulations like Europe’s Digital Services Act~\cite{de2023data}. Another approach involves using search engine result pages (SERPs) as proxies for social media data~\cite{poudel2024navigating}.

\section{Data Collection Methodology}

We compared (nearly) complete data from two social media platforms, Reddit and Twitter/X, with search engine responses for the same period.

\subsection{Reddit Data}

Reddit data was collected using the Pushshift system\footnote{\url{http://pushshift.io}} until March 2023. This dataset is comprehensive but may lack content flagged as spam by Reddit, or removed, edited, or deleted by moderators or users before collection. It also excludes content from quarantined subreddits or inaccessible posts/comments. Despite these limitations, it covers a vast majority of Reddit's visible social media content. Note that metadata such as up-/downvotes, awards, and flair may be altered post-collection and may not be fully represented in this dataset.

For this study, we focused on Reddit data from January 2023, consistent with prior research. During this period, the dataset comprised 36,090,931 posts and 253,577,506 comments across 336,949 distinct subreddits.

\subsection{X/Twitter Data}
We obtained a nearly complete X/Twitter dataset spanning 24 hours from September 20, 2022, 15:00:00 UTC, to September 21, 2022, 14:59:59 UTC using an academic API, available free at the time of collection. This dataset, though not guaranteed to be complete, aims to provide a nearly-exhaustive, stable representation of X/Twitter activity~\cite{pfeffer2023just}. During this period, 374,937,971 tweets were collected, with approximately 80\% being retweets, quotes, or replies, and the remainder original tweets.

\begin{figure*}[t!]
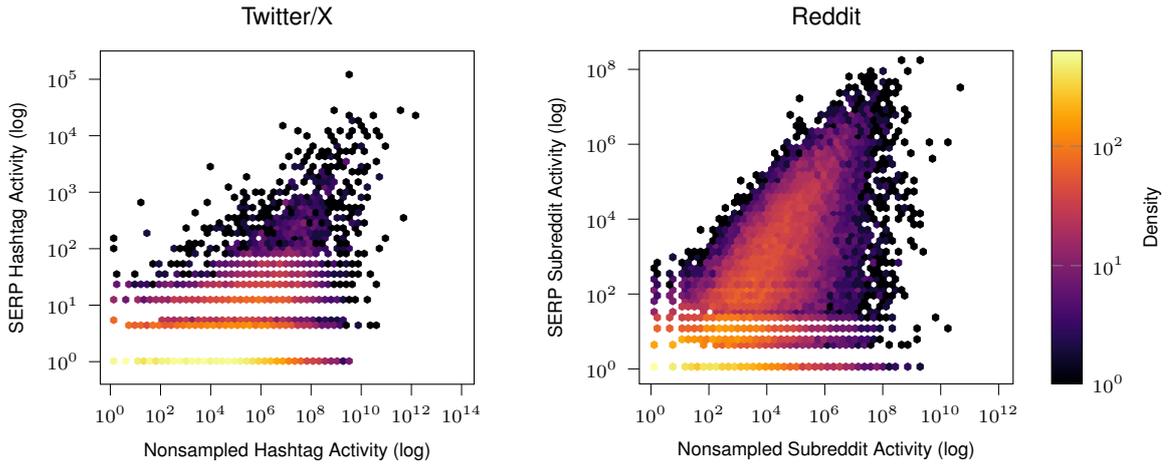

\centering
\begin{minipage}{0.4\textwidth}
  \include{test/minitwitbin}
\end{minipage}
\hspace{0.5cm}
\begin{minipage}{0.45\textwidth}
  \include{test/miniredbin}
\end{minipage}
\vspace{-1.2cm}
 \caption{Hexbin plots show correlation between hashatag and subreddit occurrence in SERP results compared to the non-sampled data for Twitter/X ($R^2=0.214$, $p<0.001)$ and for Reddit ($R^2=0.423$, $p<0.001$).} 
 \label{fig:hexbin}
\end{figure*}

\subsection{Search Engine Sampling Methodology}
Given the vast amount of social media data, extracting all indexed content from search engines is impractical. Instead, we sampled data by issuing keyword queries and extracting results from SERP. The Reddit dataset was tokenized using Lucene’s StandardAnalyzer~\cite{lucene-tokenstream}, which processes text by removing whitespace, converting to lowercase, and eliminating stopwords. We filtered tokens with non-alphabetic characters, fewer than 3 characters, or occurring less than 100 times, then a stratified sample of 1,000 keywords was selected based on document frequency for balanced representation\footnote{The list of keywords will be made available upon publication.}(see Appendix.~\ref{sample} for details). 

For each keyword, site-specific queries were issued to Google using formats like {\small\texttt{site:reddit.com \{keyword\}}} and {\small\texttt{site:twitter.com \{keyword\}}}, with time constraints set to match nonsampled Reddit data from January 2023 and Twitter/X data from September 20-21, 2022. Default query settings were maintained. The SERP-API we employed utilized multiple global proxies to mitigate geographical biases. Each query was repeated three times to account for SERP's non-deterministic nature, and results were combined across repetitions.

\subsection{Sample Statistics}

\begin{table}[t]
\centering
\caption{Number of unique subreddits and hashtags in nonsampled data and the time-matched SERP sample}
\small
\begin{tabular}{l|l|l}
\toprule
\multirow{2}{*}{\textbf{Site}} & \multicolumn{2}{c}{\textbf{Subreddits/Hashtags}} \\ \cline{2-3} 
 & \textbf{Nonsampled} & \textbf{SERP sample} \\ \hline
Reddit & 336,949 & 35,094 \\ \hline
Twitter/X & 3,014,574 & 21,920 \\ \bottomrule
\end{tabular}
\vspace{-0.5cm}
\label{tab:data-summary}
\end{table}

Relative to the enormous size of the nearly-complete Reddit and Twitter/X datasets, the time-matched SERP results yielded a total of 1,296,958 posts from Reddit and 80,651 tweets from Twitter/X. Table~\ref{tab:data-summary} shows the statistics of total unique subreddits and hashtags retrieved from the nonsampled social media data and from the SERP results for the curated list of keywords.

Rather than the posts themselves, in the present work we focus on those subreddits and hashtags returned by SERP. We conducted an in-depth comparison to understand what disparities, if any, exist between the SERP sample and the nonsampled data. This analysis is broken into four phases that correspond to the overall research questions of the present work: (1) Activity-based Analysis, (2) Characterization of the Sample, (3) Toxicity Analysis of the Sample, (4) Suppression and Promotion Analysis.

\section{Activity-based analysis}

Previous studies have shown that search engines prioritize Reddit posts with higher upvotes and tweets from users with larger followings~\cite{poudel2024navigating}. Here, we investigate whether SERP results also favor subreddits and hashtags with higher activity. We measured activity in subreddits by the number of submissions to each subreddit during the sample timeframe. Similarly, for Twitter/X, activity was measured by the frequency of each hashtag.

For Reddit, we compared the number of subreddit posts between nonsampled data and SERP samples. This comparison was visualized using hexbin plots (Fig.~\ref{fig:hexbin}), where color intensity represents data point density. On Twitter/X, we similarly compared the frequency of each hashtag between nonsampled and SERP data. Hexbin plots were chosen because they effectively display the distribution and density of large datasets, making it easier to identify patterns and correlations.


On Twitter/X, we found a moderate correlation between hashtag frequency in SERP and its occurrence in nonsampled data \textbf{($R^2=0.214$, $p<0.001$)}. For Reddit, a stronger association was observed \textbf{($R^2=0.423$, $p<0.001$)}. Interestingly, hashtags with little activity still appeared in SERP results, possibly due to sustained popularity from previous periods despite current inactivity. This trend was particularly noticeable in the Twitter/X dataset, which covers only a single day in this study.

\begin{figure}[t]
    \begin{tikzpicture}
        \begin{axis}[
            height = 3.95cm,
            width = \linewidth,
            xbar stacked,
            axis lines*=left,
            xmin=0,xmax=100,
            ytick=data,
            symbolic y coords={Not In SERP, In SERP},
            xlabel={Percentage},
            ylabel={Subreddit Type},
            tick align=outside,
            bar width=4mm,
            enlarge y limits=0.5,
            nodes near coords,
            legend style={at={(0.5,-0.25)},
            anchor=north,
            legend columns=-1,
            y tick label style={rotate=90, font=\scriptsize},
            visualization depends on={x \as \rawx},
            nodes near coords={
                \pgfmathparse{ifthenelse(\rawx>10, \rawx, "")}},
            /tikz/every even column/.append style={column sep=0.5cm}}] 
            \addplot[fill=blue!40] coordinates{(46.5,Not In SERP) (93.3,In SERP)};
            \addplot[fill=red!40] coordinates{(12.5,Not In SERP) (3.6,In SERP)};
            \addplot[fill=green!40] coordinates{(39.4,Not In SERP) (2.3,In SERP)};
            \addplot[fill=yellow!40] coordinates{(1.6,Not In SERP) (0.5,In SERP)};
            \legend{Public, Restricted, Forbidden, Private}
        \end{axis}
    \end{tikzpicture}
    \vspace{-.4cm}
     \caption{In SERP results are more likely to contain public subreddits compared to those subreddits Not In SERP results.}
    \label{fig:com_type}
\end{figure}
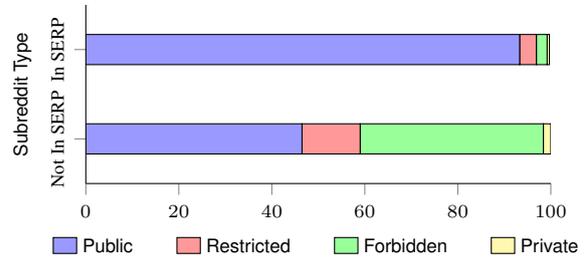

\begin{figure}[t]
    \begin{tikzpicture}
        \begin{axis}[
            ybar,
            ymin=0,
            ymax=0.5,
            width=5.5cm,
            height=4cm,
            bar width=6pt,
            symbolic x coords={Other, Entertainment, Games, Advertisement, Finance, Sex, Politics, Celebrities, Unknown},
            xtick=data,
            ylabel={Proportion},
            xlabel={},
            width=0.5\textwidth,
            x tick label style={rotate=45,anchor=east, font=\scriptsize},
            ]
            \addplot[
            pattern=horizontal lines, 
            pattern color=blue] coordinates {
            (Other,0.300)(Entertainment,0.2975) (Games,0.105) (Advertisement,0.085) (Finance,0.04) (Sex,0.015) (Politics,0.074) (Celebrities,0.0625) (Unknown,0.025)};
            \addplot+[
            pattern=dots, 
            pattern color=red] coordinates {
            (Other,0.425)(Entertainment,0.1875) (Games,0.145) (Advertisement,0.055) (Finance,0.06) (Sex,0.055) (Politics,0.035) (Celebrities,0.035) (Unknown,0.0025)};
            \legend{Not In SERP, In SERP}
        \end{axis}
    \end{tikzpicture}
    \caption{Distribution of the hashtag categories for those found In SERP results compared to those Not In SERP results.}
    \label{fig:gpt}
\end{figure}
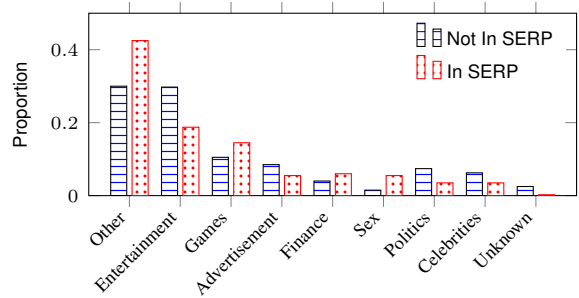

\subsection{Characterization of Sampled Subreddits and Hashtags}
\label{sec:char}

Our analysis showed a moderate correlation between subreddit and hashtag engagement and SERP visibility. Here, we explore deeper by examining which types of subreddits and hashtags are overrepresented or underrepresented in SERP compared to an unbiased sample of the data. Specifically, we focus on the top 1,000 most active subreddits and English hashtags based on post frequency on Reddit and Twitter/X, respectively.

On Reddit, subreddits are categorized into following visibility states: public, restricted, forbidden (banned by Reddit as of March 2023), or private (visible only to subscribed members). Our analysis shows that SERP significantly favors public subreddits and suppresses those categorized as restricted, forbidden, and private; Fig.~\ref{fig:com_type} illustrates the proportions of subreddit types returned and not returned by SERP. Using OpenAI's GPT-4~\cite{kublik2023gpt}, we categorized each Twitter/X hashtag into one of nine previously identified categories~\cite{pfeffer2023just}, as shown in Fig.~\ref{fig:gpt}. The prompt template is shown in Appendix~\ref{sec:prompt}. On SERP, categories like Games and Finance were over-represented, while Advertisement, Politics, and Entertainment were under-represented compared to the 'Not in SERP' category. These findings are specific to the hashtags prevalent during a 24-hour period in late September 2022 and may not reflect broader trends on Twitter/X. (See Appendix (Tables.~\ref{tab:subreddit_cat} \& ~\ref{tab:hashtag_cat}) for some of the representative subreddits/hashtags within each categories/classes respectively.)

Next, we will analyze the content within these top 1000 subreddits and hashtags, examining the types of posts appearing in SERP versus those that do not, using a toxicity analysis.

\subsection{Toxicity Analysis}

Toxicity in online communities is a critical research area requiring complete social media data access. It's vital to determine if SERP-represented groups truly reflect overall toxicity dynamics.
Traditional toxicity analysis relied on keyword presence for identifying toxic content~\cite{rezvan2020analyzing}. Transformer models like BERT now lead, adapting to evolving cultural and linguistic contexts~\cite{devlin2018bert,sheth2022defining}. We employed Toxic-BERT~\cite{Detoxify}, trained on annotated Wikipedia comments, to assess toxicity in Reddit post titles and Tweets. It provides probabilities for toxicity, obscenity, and insults, with other labels (threat, severe\_toxic, identity\_hate) being extremely rare and not shown in our results. We compared the toxicity levels across two categories: \textit{In SERP} and \textit{Not In SERP}. The "In SERP" group consists of randomly sampled 5,000 posts that appeared directly in search engine results, specifically within the top 1,000 results for selected subreddits and hashtags. The "Not In SERP" group includes 5,000 posts randomly selected from subreddits and hashtags not indexed by search engines, ensuring that none of these posts were visible in search results.

By comparing these samples, we assessed and contrasted toxicity levels among posts from subreddits and hashtags that are in SERP, and not in SERP . This helps us understand how search engine indexing and result presentation might influence users' exposure to toxic content.

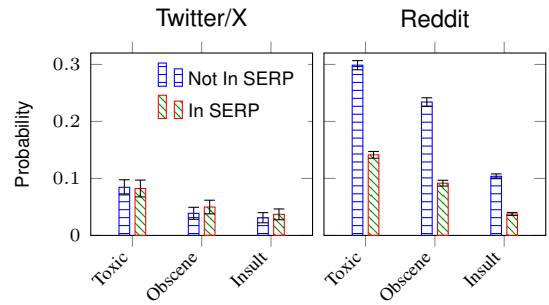
\begin{figure}[t]
    \begin{tikzpicture}
    \begin{groupplot}[group style={group size= 2 by 1, horizontal sep=0.15cm},
    height=4cm,width=4.5cm,
    ymin=0,
    ymax=0.32,
    xlabel={},
    xtick=data,
    ylabel={},    
    xticklabels={Toxic, Obscene, Insult},
    x tick label style={rotate=45,anchor=east, font=\scriptsize},
    enlarge x limits=0.3,
    error bars/.cd,
    y dir=normal,    
    ]
        \nextgroupplot[
            title={Twitter/X},
            ylabel={Probability},
            ybar, 
            bar width=4pt, 
            legend entries = {Not In SERP, In SERP},
        ]
        
        \addplot+[
            pattern=horizontal lines, 
            pattern color=blue, 
            error bars/.cd,
            y dir=both,
            y explicit,
            error bar style={black},
        ] coordinates {
            (1, 0.08436567521974211) +- (0, 0.013298577540542726)
            (2, 0.03877794858251582) +- (0, 0.01053535507064628)
            (3, 0.031070066814951135) +- (0, 0.00889745029117596)
        };
        
        
        \addplot+[
            pattern=north west lines, 
            pattern color=green!60!black, 
            error bars/.cd,
            y dir=both,
            y explicit,
            error bar style={black},
        ]coordinates {
            (1, 0.08231162735581166) +- (0, 0.014732396794840074)
            (2,0.04981125289767806) +- (0, 0.011939922913473636)
            (3,0.03696752363684937) +- (0, 0.009482830939231844)
        };

\nextgroupplot[
            ybar, 
            bar width=4pt, 
            yticklabels={},
            title={Reddit},            
        ]
        
        \addplot+[
            pattern=horizontal lines, 
            pattern color=blue, 
            error bars/.cd,
            y dir=both,
            y explicit,
            error bar style={black},
        ]coordinates {
            (1, 0.2986632363237091) +- (0, 0.008011179168351402)
            (2, 0.23402565614601625) +- (0, 0.007445270345708586)
            (3, 0.10371601037359797) +- (0, 0.004222221621008083)
        };

        
        \addplot+[
            pattern=north west lines, 
            pattern color=green!60!black, 
            error bars/.cd,
            y dir=both,
            y explicit,
            error bar style={black}, 
        ] coordinates {
            (1, 0.14133806869261317) +- (0, 0.0059441528076461)
            (2, 0.09165523601786117) +- (0, 0.0050853470557816515)
            (3, 0.03771301613801479) +- (0, 0.0025815950862881115)
        };
        
        \end{groupplot}
        
    \end{tikzpicture}
  \vspace{-1.1cm}
  \caption{Toxicity analysis of subreddits and hashtags from Reddit and Twitter/X resp. demonstrates that subreddits and hashtags returned exclusively by the SERP are less likely to be toxic compared to those not included in the SERP.}
  \label{fig:toxic}
\end{figure}


Figure~\ref{fig:toxic} illustrates the mean label probabilities alongside their 95\% confidence intervals, highlighting key differences between Reddit and Twitter/X in terms of content toxicity. Our analysis reveals mixed results. Subreddits that do not appear in SERP exhibited higher toxicity levels compared to those that do appear or are returned by SERP suggesting that SERP aggressively filters subreddits. On Twitter/X, hashtags Not In SERP were only marginally more toxic than those In SERP, showing little difference overall. These findings may reflect the content landscape of Twitter/X during the time of data collection, where prominent discussions focused on less controversial topics, such as entertainment, finance, gaming, and current events.

Despite these platform-specific variations, the overall toxicity of Twitter/X content was lower than that of Reddit. This may be attributed to Reddit's higher prevalence of subreddits focused on adult content, which tend to be perceived as more toxic. However, as shown in Figure~\ref{fig:com_type}, such subreddits represent only a small subset of the most popular communities on Reddit.

\section{Suppression and Promotion}
\label{sec:inclintion}

While the previous categorization sheds light on the types and nature of subreddits and hashtags retrieved by SERP, it overlooks how frequently they appear, potentially introducing bias in their portrayal compared to nonsampled data. In this section, we treat subreddits and hashtags as tokens and employ conventional token analysis to assess their suppression and promotion in SERP. Various statistical analyses can be used to compare these distributions~\cite{cha2007comprehensive,deza2006dictionary}. However, traditional methods face challenges with Zipfian data typical of most text datasets~\cite{gerlach2016similarity,dodds2023allotaxonometry}. To address this, we utilize Rank Turbulence Divergence (RTD)~\cite{dodds2023allotaxonometry} to quantify the disparity between the activity distribution of nonsampled subreddits and hashtags and those retrieved in the SERP sample; see Appendix.~\ref{sec:rtd_eq} for details. 

A lower score indicates low rank divergence, indicating similar distributions. Conversely, a higher score suggests larger divergence. Table \ref{tab:rtd} shows the mean RTD for SERP results compared to non-sampled social media data across all 1,000 keywords, highlighting significant disparities in this domain-level analysis\footnote{A control test found an RTD of \~0.30 for random comparisons within the Reddit/X dataset~\cite{poudel2024navigating}}.

\begin{table}[t]
    \centering
    \caption{Rank Turbulence Divergence (RTD) between SERP subreddits and hashtags and the nonsampled social media subreddits and hashtags.}
    \vspace{-.1cm}
    \small
    \begin{tabular}{lc}
    \toprule
        \textbf{Site} & \textbf{RTD} (SERP Sample vs Nonsampled)\\ \midrule
        Reddit & 0.64 \\
        Twitter/X & 0.73 \\ \bottomrule
    \end{tabular}
    \vspace{-0.2cm}
    \label{tab:rtd}
\end{table}

\begin{figure*}[t]
\begin{tikzpicture}
\begin{axis}[
    height = 6cm,
    width=.6\textwidth,
    ymin=-0.7,
    ymax=0.9,
    ytick={-0.6,-0.4,-0.2,0,0.2,0.4,0.6,0.8},
    xmin=10, 
    xmax=1000000, 
    xmode=log, 
    axis lines=left,
    xlabel={Subreddit Activity (log)},
    ylabel={Reddit RTD},
]
\node [above right, inner sep=0pt] at (11, -0.69){\includegraphics[width=.43\textwidth,keepaspectratio]{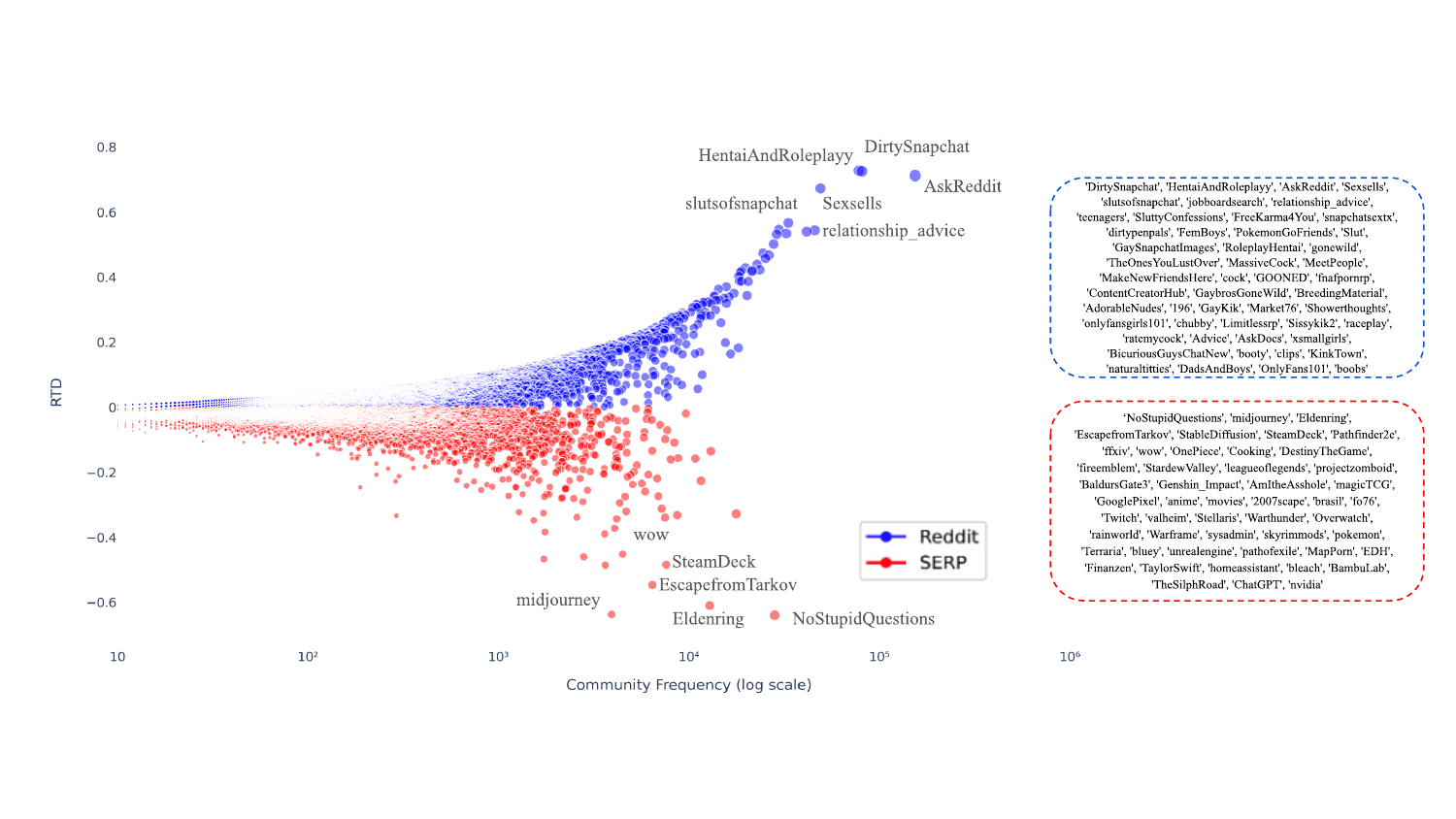}};
\end{axis}
  \node [above right, inner sep=0pt] at (8.75, 0.0){ 
    \scriptsize
    \begin{tabular}{ll}
    \multicolumn{2}{c}{Most Divergent Subreddits} \\
    \toprule
  \textcolor{blue}{Reddit} & \textcolor{red}{SERP} \\ \midrule
    \textcolor{blue}{/r/slutsofsnapchat} & \textcolor{red}{/r/StableDiffusion} \\
    \textcolor{blue}{/r/jobboardsearch} & \textcolor{red}{/r/Pathfinder2e} \\
    \textcolor{blue}{/r/teenagers} & \textcolor{red}{/r/ffxiv} \\
    \textcolor{blue}{/r/SluttyConfessions} & \textcolor{red}{/r/OnePiece} \\
    \textcolor{blue}{/r/FreeKarma4You} & \textcolor{red}{/r/Cooking} \\
    \textcolor{blue}{/r/snapchatsextx} & \textcolor{red}{/r/DestinyTheGame} \\
    \textcolor{blue}{/r/dirtypenpals} & \textcolor{red}{/r/fireemblem} \\
    \textcolor{blue}{/r/FemBoys} & \textcolor{red}{/r/StardewValley} \\
    \textcolor{blue}{/r/PokemonGoFriends} & \textcolor{red}{/r/leagueoflegends} \\
    \textcolor{blue}{/r/Slut} & \textcolor{red}{/r/projectzomboid} \\
    \textcolor{blue}{/r/GaySnapchatImages} & \textcolor{red}{/r/BaldursGate3} \\
    \textcolor{blue}{/r/RoleplayHentai} & \textcolor{red}{/r/Genshin\_Impact} \\
    \textcolor{blue}{/r/gonewild} & \textcolor{red}{/r/AmItheAsshole} \\
    \textcolor{blue}{/r/TheOnesYouLustOver} & \textcolor{red}{/r/magicTCG} \\ \bottomrule
      \end{tabular}
  };
  \end{tikzpicture}
  \begin{tikzpicture}
\begin{axis}[
    height = 6cm,
    width=.6\textwidth,
    ymin=-0.9,
    ymax=0.9,
    ytick={-0.8,-0.6,-0.4,-0.2,0,0.2,0.4,0.6,0.8},
    xmin=10, 
    xmax=10000000, 
    xmode=log, 
    axis lines=left,
    xlabel={Hashtag Activity (log)},
    ylabel={Twitter/X RTD},
]
\node [above right, inner sep=0pt] at (11, -0.817){\includegraphics[width=.45\textwidth,keepaspectratio]{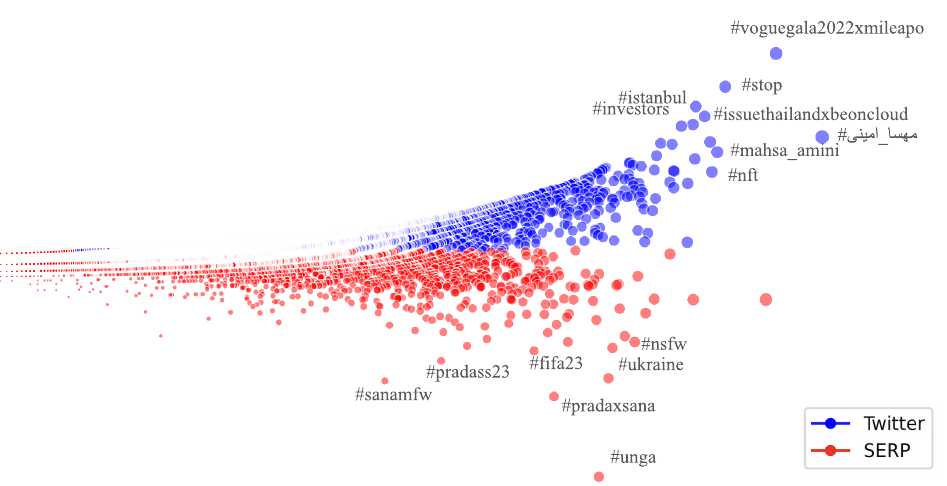}};
\end{axis}
  \hfill
  \node [above right, inner sep=0pt] at (8.75, 0.0){  
    \scriptsize
    \begin{tabular}{ll}
    \multicolumn{2}{c}{Most Divergent Hashtags} \\ \toprule
  \textcolor{blue}{Twitter/X} & \textcolor{red}{SERP} \\ \midrule
    \textcolor{blue}{\#bifw2022xbeoncloud} & \textcolor{red}{\#unga77} \\
    \textcolor{blue}{\#enhypen} & \textcolor{red}{\#russia} \\
    \textcolor{blue}{\#trumpmeltdown} & \textcolor{red}{\#firstdayoffall} \\
    \textcolor{blue}{\#putinisdead} & \textcolor{red}{\#cybersecurity} \\
    \textcolor{blue}{\#milephakphum} & \textcolor{red}{\#linux} \\
    \textcolor{blue}{\#biblebuild} & \textcolor{red}{\#abbottelementary} \\
    \textcolor{blue}{\#eth} & \textcolor{red}{\#hijab} \\
    \textcolor{blue}{\#nftday} & \textcolor{red}{\#peaceday} \\
    \textcolor{blue}{\#trudeaumustgo} & \textcolor{red}{\#worldalzheimersday} \\
    \textcolor{blue}{\#trump} & \textcolor{red}{\#pakveng} \\
    \textcolor{blue}{\#tigraygenocide} & \textcolor{red}{\#bachelorvote} \\
    \textcolor{blue}{\#cryptocurrency} & \textcolor{red}{\#onlyfans} \\
    \textcolor{blue}{\#iranianlivesmatter} & \textcolor{red}{\#worldrhinoday} \\
    \textcolor{blue}{\#kashmir} & \textcolor{red}{\#money} \\ \bottomrule
      \end{tabular}
      };
      \end{tikzpicture}
  \caption{Rank Turbulence Divergence (RTD) of ranked Subreddits and Hashtags as a function of activity. Subreddits and hashtags with higher likelihood in nonsampled social media data are represented in \textcolor{blue}{blue}, while those with higher likelihood in SERP results are in \textcolor{red}{red}.}
  \label{fig:macroreddit}
\end{figure*}

\subsection{Divergence versus Frequency}
Selecting on only those subreddits and hashtags that appeared at least once in SERP results, we characterized their inclinations, \ie, if the subreddit is more or less likely to appear in the SERP sample compared to the non sampled social media data, and plotted these signed divergences as a function of the activity. Figure~\ref{fig:macroreddit} illustrates the most divergent subreddits (top) and hashtags (bottom). Additionally, Fig~\ref{fig:commrtdreddit} in the Appendix shows the distributions of the 15 highest and lowest individual divergences (Eq. \ref{inverse_control}) and their mean (representing Eq. \ref{rtd}) for each subreddit and hashtag respectively.

For Twitter/X hashtags, SERP prominently featured hashtags related to events like the United Nations General Assembly (UNGA), the FIFA video game, and hashtags about the fashion-house Prada and its appearance at Milan Fashion Week (MFW). These events occurred during or prior to the data collection period. On the contrary, hashtags related to the appearance of two Thai celebrities Mile and Apo at the Vogue Gala as well as their talent agency BeOnCloud were largely hidden from SERP results. A hashtag of Mahsa Amini, an Iranian woman who refused to wear a headscarf and died under suspicious circumstances in the days prior to data collection was also comparatively hidden from SERP results. Cryptocurrency hashtags related to investors and NFTs were comparatively hidden from SERP results as well. Most common hashtags from each inclination are listed on the right.
Similarly, for Reddit, as demonstrated in the previous analysis, gaming and conversational subreddits are more frequently returned in SERP results, while subreddits focused on adult content are more prevalent on Reddit. Interestingly, /r/AskReddit and /r/relationship\_advice are notably less visible in SERP results, and requires a further exploration.

\subsection{Coverage of Subreddits}

\begin{figure}[h]
    \includegraphics[width=\linewidth]{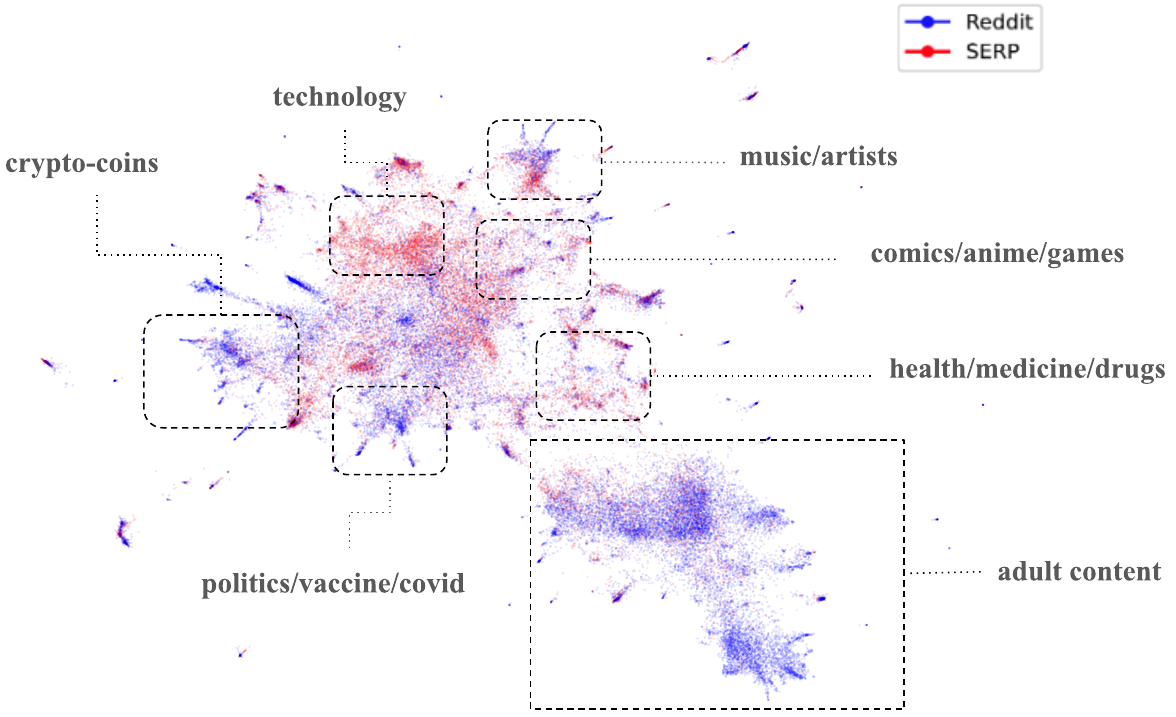}
    \caption{Semantic embeddings of subreddits that are found In SERP (\textcolor{red}{red}) and Not In SERP (\textcolor{blue}{blue}). Clusters of subreddits about adult content, political, crypto-coins are generally absent from SERP results.}
    \label{fig:map}
\end{figure}

We conducted a case study comparing subreddits included in SERP with those not included, as illustrated in Fig.\ref{fig:map}. For each subreddit with at least 10 posts, we semantically mapped the content using MPNet-Base-V2 embeddings, averaged from five random posts per subreddit. We then used UMap to project these embeddings into a two-dimensional space~\cite{mcinnes2018umap}.

Red points denote subreddits in SERP, while blue points denote those not in SERP. We identified seven clusters, where clusters dominated by red or blue indicate SERP status. Pornographic and adult content was notably absent from SERP, while technology, music, comics, games, and health-related subreddits were prominently featured. Conversely, subreddits discussing crypto-coins, politics, and COVID-19 were less likely to appear in SERP.

\section{Discussion}
Our study demonstrates how search engines act as gatekeepers, shaping online discourse by selectively surfacing a biased subset of subreddits and hashtags in their SERPs. This selective visibility directly impacts how users access and engage with information. By analyzing the patterns of inclusion and exclusion within SERP results, we observe how search engine algorithms and moderation practices play a central role in framing the topics and communities that dominate online conversations.

We found that subreddits and hashtags with higher engagement levels, such as highly upvoted Reddit posts or popular hashtags, are more likely to appear in SERPs. This tendency was more pronounced on Reddit, where there is a stronger correlation between engagement metrics and SERP appearance compared to X/Twitter. This disparity suggests that search engine algorithms treat engagement metrics differently across platforms, reflecting the unique dynamics of user interactions on each. This insight directly supports our first research question by revealing the role of search engine algorithms in amplifying content with higher activity and participation.


Of course, the time-scope of our data revealed specific events, such as \#climateweeknyc and \#nationalfitnessday. These hashtags gained prominence around their corresponding events, indicating that search engines respond to temporal spikes in engagement and act as curators of public discourse.

Notably, political subreddits and hashtags were systematically less likely to appear in SERP, suggesting that factors beyond user engagement, such as moderation policies and content restrictions, significantly influence visibility. Political content, along with discussions related to pornography, bots, and cryptocurrency, were disproportionately filtered from SERPs. This underscores the gatekeeping function of search engines, which, through moderation, both maintain the quality of the content they display and inadvertently suppress discourse in these areas.

Our analysis shows that SERPs filter out content related to pornography, bots, and cryptocurrency, likely due to moderation policies aimed at reducing inappropriate content. While this helps create a safer online space, it also suppresses legitimate discussions, skewing the available discourse. 

The toxicity analysis adds an important dimension to these findings. We found that content surfaced by SERPs generally contains less toxic language compared to the content from subreddits and hashtags that do not appear in SERP results. This suggests that search engines are effectively reducing exposure to harmful or toxic content. While this can be seen as a positive step towards creating a safer and more civil online environment, it also introduces concerns about over-filtering. Specifically, by aggressively limiting toxic content, search engines might also suppress important discussions that could be critical to public discourse. This observation directly informs our third research question by showing how moderation policies tangibly shape the nature of the content users access, and raising questions about the balance between safety and free expression.

The results of our study have several implications. They suggest that SERP algorithms and moderation policies collectively shape the online information landscape in ways that may not be immediately apparent to users. By favoring certain communities and suppressing others, SERPs can influence public discourse, access to information, and the diversity of viewpoints available to users.

\subsection{Conclusions}

In conclusion, our study highlights the significant role that SERP rankings and moderation play in shaping the visibility and representation of online communities. By bringing attention to the biases inherent in SERP results, we hope to encourage further investigation and dialogue on how to promote a more inclusive and representative online environment. Future research should explore the inner workings of SERP algorithms and moderation policies to understand the criteria that drive content prioritization and filtering.  Additionally, investigating the broader societal impact—such as how user behavior and trust in online information are shaped by SERP biases—could provide deeper insights into how search engines influence public perceptions and interactions with digital content.

Expanding this research to other platforms and search engines will be crucial for determining whether similar biases are prevalent elsewhere. Further, longitudinal studies could track how these dynamics change over time in response to shifts in moderation practices, algorithm updates, or public sentiment. Understanding the long-term effects of these biases will be key to informing policies that ensure both the integrity of public discourse and the promotion of a more inclusive online environment.


\subsection{Limitations}

This study is not without limitation. First, the non-deterministic nature of the SERP API means that the collected data represents only a sample of the search engine output, which could affect frequency-based analysis. To mitigate potential variations, we employed different data collection strategies: for Reddit, data was collected daily for each keyword, and for Twitter/X, the SERP queries were run three times per keyword. Although keyword choice might influence results, our five-fold cross-validation, which analyzed five different 80\% samples of the keyword list, yielded similar results, giving confidence that our findings are robust.

Another limitation arises due to the difference in data coverage between Twitter/X and Reddit. Although the Twitter dataset is complete coverage for a single day, the Reddit data spans a full month. This variance in data completeness may introduce inconsistency in the generalization of the findings between these sites. However, considering that our Twitter/X dataset represents the only complete dataset currently accessible for research purposes, we remain optimistic that our conclusions are more generalizable than any other methodology or dataset available.

Finally, we focused primarily on English hashtags for this study, which required filtering out many other hashtags. While this approach may have resulted in a loss of information and missed findings, it was a deliberate and necessary decision to maintain consistency across the experiments.

\section*{Acknowledgments}
We would like to thank Adnan Hoq for his helpful discussion. This research is sponsored by the University of Notre Dame Democracy Initiative.

\newpage

\appendix

\renewcommand{\thefigure}{A\arabic{figure}}
\setcounter{figure}{0}

\section{Appendix}
\label{appendix}

\subsection{Keyword Sampling}
\label{sample}

To construct a representative keyword sample, we employed stratified sampling based on word frequency, addressing the Zipfian distribution of language usage, where a small subset of terms is highly frequent while the majority are rare. Random sampling alone would disproportionately underrepresent frequent terms. In our approach, keywords were sorted by frequency, and every \(N\)-th term was selected (\(N = \text{int}(1000/\text{num\_terms})\)), ensuring a balanced inclusion of common, medium, and rare terms.  

In large-scale language usage (e.g., billions of tokens), token and term distributions converge, making this stratified sampling generalizable across platforms. This assumption is supported by empirical evidence: a recent study found a strong \(R = 0.95\) correlation between term distributions on Facebook and Twitter~\cite{herdaugdelen2017social}. Similarly, research comparing social media comments and tweets observed “relatively high similarity” in lexical distributions~\cite{baldwin2013noisy}. While our analysis focuses on Reddit and Twitter, these findings suggest comparable cross-platform distributional properties.  

The stratified sampling method naturally includes common terms \textit{(e.g., "like", "first", "year")} that are frequently used across platforms, making the sample well-suited for cross-platform studies. This balanced approach ensures that the keyword set captures both platform-specific and shared language patterns, supporting robust comparisons across diverse contexts.

\subsection{Distribution of Hashtags}

\subsubsection{Prompting Template for Hashtags Classification}
\label{sec:prompt}

\noindent \texttt{instruction: Classify the given hashtag into one of the following topics: [games, politics, celebrities, sex, entertainment, advertisement, finance, Unknown, other]. Choose only one, and provide the topic only.} \\

\noindent \texttt{Instruction: [instruction]\\
Hashtag: \{hashtag\}} \\

\noindent In the prompt, the hashtag is selected from the set of Top 1000 hashtags, i.e., $hashtag \sim \{\text{Top 1000 hashtags}\}$.


\renewcommand\theequation{E.\arabic{equation}}

\subsection{Rank Turbulence Divergence (RTD)}
\label{sec:rtd_eq}
Formally, let R1 and R2 be two distributions ranked from most active to least active. Initially, the RTD computes the element-wise divergence through the following process:

\begin{equation}
\left|\frac{1}{[r_{\xi, 1}]^\alpha}- \frac{1}{[r_{\xi,2}]^\alpha}\right|^{\frac{1}{\alpha+1}}
\label{inverse_control}
\end{equation}
\noindent where $\xi$ represents a token (\ie, subreddit or hashtag) and $r_{\xi, 1}$ and $r_{\xi, 2}$ denote its ranks within R1 and R2, respectively and a control parameter $\alpha$ that regulates the importance of rank. For each token present in the combined domain of R1 and R2, we compute their divergence using Eq.~\ref{inverse_control}. In the present work, we use $\alpha = \frac{1}{3}$, which has been shown in previous work to deliver a reasonably balanced list of words with ranks from across the common-to-rare spectrum~\cite{dodds2023allotaxonometry}.

The final RTD is a comparison of R1 and R2 summed over the element-level divergence. It includes a normalization prefactor $N_{1,2;\alpha}$ and takes the following form. 

\begin{equation}
\begin{aligned}
&RTD^R_\alpha(R1 \parallel R2)  \\
&= \frac{1}{N_{1,2;\alpha}} \frac{\alpha+1}{\alpha} \sum\limits_{\xi \in R_{1,2; \alpha}}
\left|\frac{1}{[r_{\xi, 1}]^\alpha} - \frac{1}{[r_{\xi, 2}]^\alpha}\right|^{\frac{1}{\alpha+1}}
\end{aligned}
\label{rtd}
\end{equation}

\setcounter{table}{0}
\renewcommand{\thetable}{T\arabic{table}}

\subsection{Hashtags/Subreddits by Categories}
Tables~\ref{tab:subreddit_cat} and~\ref{tab:hashtag_cat} present a selection of subreddits and hashtags respectively, based on their visibility in Search Engine Results Pages (SERP) \ie \textit{'In SERP'} and \textit{'Not In SERP"}. 

\begin{table}[h]
\caption{Selected representative subreddits by their Reddit visibility designation}
\label{tab:subreddit_cat}
\setlength{\tabcolsep}{3pt}
\small{
\begin{tabular}{lll}
\toprule
&\multicolumn{1}{c}{\textbf{In SERP}}   & \multicolumn{1}{c}{\textbf{Not in SERP}} \\
\midrule
\multirow{5}{*}{\rotatebox[]{90}{Public}}
& /r/AskReddit          & /r/dirtyr4r             \\
& /r/HentaiAndRoleplayy & /r/rapefantasies        \\ 
& /r/DirtySnapchat      & /r/SchoolgirlsXXX              \\
& /r/presonalfinance    & /r/StockTradingIdeas    \\ 
& /r/pokemon            & /r/CamSluts             \\ 
\midrule

\multirow{5}{*}{\rotatebox[]{90}{Restricted}}
& /r/AndrewTateTop         & /r/AutoNewspaper        \\ 
& /r/DeathObituaries    & /r/DenverhookupF4M      \\
& /r/DemocraticUnderground & /r/NaughtyRealGirls     \\ 
& /r/BustyNaturals         & /r/CoinMarketDo         \\
& /r/NaughtyWives          & /r/rice\_cakes          \\
\midrule

\multirow{5}{*}{\rotatebox[]{90}{Forbidden}}
& /r/RoleplayHentai       & /r/SextOnSnapchat         \\
& /r/GOONED               & /r/nflstreamlinks         \\
& /r/GayKik               & /r/nudecutegirls          \\
& /r/PussyFlashing        & /r/SATXhot\_momsNwives    \\
& /r/ContentCreatorHub    & /r/horny \\      
\midrule

\multirow{4}{*}{\rotatebox[]{90}{Private}}
& /r/FIFA                & /r/N\_E\_W\_S         \\
& /r/NSFW\_Social        & /r/Pennsylvaniaswingersr         \\
& /r/Balls               & /r/northwestohiohookups         \\
& /r/MeetPeople          & /r/IndiaOpen              \\
\bottomrule
\end{tabular}
}
\end{table}


\begin{table}[t]
\caption{Selected representative hashtags by their classification}
\label{tab:hashtag_cat}
\setlength{\tabcolsep}{3pt}
\small{
\begin{tabular}{lll}
\toprule
&\multicolumn{1}{c}{\textbf{In SERP}}   & \multicolumn{1}{c}{\textbf{Not in SERP}} \\
\midrule
\multirow{5}{*}{\rotatebox[]{90}{Other}}&\#saveocws     & \#weathercloud             \\
&\#worldalzheimersday        & \#authorsoftwitter        \\ 
&\#worldpeaceday     & \#christianity               \\
&\#nationalfitnessday      & \#snapchatleak    \\ 
&\#climateweeknyc       & \#the\_golden\_hour             \\ 
\midrule

\multirow{5}{*}{\rotatebox[]{90}{Entertainment}}&\#issuethailandxbeoncloud     & \#bkppthedocumentary             \\
&\#houseofthedragon        & \#shadowhunters        \\ 
&\#thebachelorette     & \#bigmouth               \\
&\#manifesto\_in\_seoul      & \#houseofthedragonhbo   \\ 
&\#thevoice       & \#tohseason3             \\ 
\midrule

\multirow{5}{*}{\rotatebox[]{90}{Games}}&\#fortniteart     & \#fortnitechapter3season4             \\
&\#genshinimapct        & \#genshingiveaway        \\ 
&\#sonicthehedgehog     & \#twitchfr               \\
&\#playstation      & \#phiballs    \\ 
&\#fifaworldcup       & \#battleship             \\ 
\midrule

\multirow{5}{*}{\rotatebox[]{90}{Advertisement}}&\#bifw2022xbeoncloud     & \#nftgiveway             \\
&\#shopmycloset        & \#chimepridepayssweeps        \\ 
&\#buyingcontent     & \#followback               \\
&\#cashappboostweek      & \#tlp\_promotion    \\ 
&\#iphone13onflipkart       & \#earlyaccessisliveonmyntra             \\ 
\midrule

\multirow{5}{*}{\rotatebox[]{90}{Political}}&\#government  & \#forabolsonaro       \\ 
&\#putinwarcriminal    & \#seditionhunters       \\ 
&\#fbimostwanted       & \#toriesout75       \\ 
&\#standwithukraine       & \#trumprally       \\ 
&\#freepalestine       & \#stopgopabortionbans       \\ 
\bottomrule
\end{tabular}
}
\end{table}

\newpage
\subsection{Divergence Versus Frequency}

\begin{figure*}
\centering
\begin{minipage}{\textwidth}
\begin{minipage}{.45\textwidth}
    \pgfplotstableread[col sep= tab]{box/twitter_top.csv}\datatable
\pgfplotstableread[col sep=tab]{box/twitter_bot.csv}\databottable
\begin{tikzpicture}
    \begin{axis}[
        name=Ax1,
        width=.84\linewidth,
        height=5cm, 
        xlabel={},
        title={\textcolor{black}{Twitter/X}},
        xtick=\empty,
        yticklabels from table={\databottable}{x},
        ytick = {1,...,26},
        ylabel={},
        enlarge y limits={abs=2pt},
        xmin=-0.5,xmax=0.5,
        ymajorgrids,
        red,
        yticklabel style={font=\scriptsize, black},
        axis line style={black},
        boxplot/draw direction=x,
        boxplot/average={auto},
        boxplot/every average/.style={/tikz/mark=*,mark size=1},
    ]
\boxplotprepared{\databottable}{\#malta}
\boxplotprepared{\databottable}{\#survey}
\boxplotprepared{\databottable}{\#discrimination}
\boxplotprepared{\databottable}{\#flood}
\boxplotprepared{\databottable}{\#transrightsarehumanrights}
\boxplotprepared{\databottable}{\#staytuned}
\boxplotprepared{\databottable}{\#lbc}
\boxplotprepared{\databottable}{\#npaw2022}
\boxplotprepared{\databottable}{\#newstoday}
\boxplotprepared{\databottable}{\#phdchat}
\boxplotprepared{\databottable}{\#ink}
\boxplotprepared{\databottable}{\#justsayin}
\boxplotprepared{\databottable}{\#supremecourtofindia}
\boxplotprepared{\databottable}{\#nationsleague}
\boxplotprepared{\databottable}{\#aewdynamite}
    \end{axis}
    \begin{axis}[
        at={($(Ax1.south east)+(0,-0.25cm)$)},anchor=north east,
        width=.84\linewidth,
        height=5cm, 
        xlabel style = {align=center},
        xlabel={Signed RTD \\ \textcolor{red}{$\xleftarrow{\textrm{More Likely Google}}$}\hspace{1cm}\textcolor{blue}{$\xrightarrow{\textrm{More Likely Twitter/X}}$}},        
        yticklabels from table={\datatable}{x},
        ytick = {1,...,26},
        ylabel={},
        xmin=-0.5,xmax=0.5,
        enlarge y limits={abs=2pt},
        ymajorgrids,
        blue,
        yticklabel style={font=\scriptsize, black},
        every axis label/.append style ={black},
        every tick label/.append style={black},       
        axis line style={black},        
        boxplot/draw direction=x,
        boxplot/average={auto},
        boxplot/every average/.style={/tikz/mark=*,mark size=1},
    ]
\boxplotprepared{\datatable}{\#voguegala2022xmileapo}
\boxplotprepared{\datatable}{\#nft}
\boxplotprepared{\datatable}{\#nsfwtwt}
\boxplotprepared{\datatable}{\#ak61}
\boxplotprepared{\datatable}{\#munawarfaruqui}
\boxplotprepared{\datatable}{\#crypto}
\boxplotprepared{\datatable}{\#nowplaying}
\boxplotprepared{\datatable}{\#nfts}
\boxplotprepared{\datatable}{\#stopwarontigray}
\boxplotprepared{\datatable}{\#godmorningwednesday}
\boxplotprepared{\datatable}{\#aliexpress}
\boxplotprepared{\datatable}{\#bitcoin}
\boxplotprepared{\datatable}{\#rbxs}
\boxplotprepared{\datatable}{\#mahsaamini}
\boxplotprepared{\datatable}{\#bbnaija}
    \end{axis}    
\end{tikzpicture}
\end{minipage}
    \hfill
\begin{minipage}{.45\textwidth}    
    \pgfplotstableread[col sep=tab]{box/reddit_top.csv}\datatable
\pgfplotstableread[col sep=tab]{box/reddit_bot.csv}\databottable
\begin{tikzpicture}
    \begin{axis}[
        name=Ax1,
        width=.84\linewidth,
        height=5cm, 
        xlabel={},
        title={\textcolor{black}{Reddit}},
        xtick=\empty,
        yticklabels from table={\databottable}{x},
        ytick = {1,...,26},
        ylabel={},
        enlarge y limits={abs=2pt},
        xmin=-0.5,xmax=0.5,
        ymajorgrids,
        red,
        yticklabel style={font=\scriptsize, black},
        axis line style={black},
        boxplot/draw direction=x,
        boxplot/average={auto},
        boxplot/every average/.style={/tikz/mark=*,mark size=1},
    ]
\boxplotprepared{\databottable}{Eldenring}
\boxplotprepared{\databottable}{StardewValley}
\boxplotprepared{\databottable}{bleach}
\boxplotprepared{\databottable}{sysadmin}
\boxplotprepared{\databottable}{skyrimmods}
\boxplotprepared{\databottable}{StableDiffusion}
\boxplotprepared{\databottable}{DestinyTheGame}
\boxplotprepared{\databottable}{projectzomboid}
\boxplotprepared{\databottable}{SteamDeck}
\boxplotprepared{\databottable}{EscapefromTarkov}
\boxplotprepared{\databottable}{fireemblem}
\boxplotprepared{\databottable}{wow}
\boxplotprepared{\databottable}{Pathfinder2e}
\boxplotprepared{\databottable}{BaldursGate3}
\boxplotprepared{\databottable}{ffxiv}
    \end{axis}
    \begin{axis}[
        at={($(Ax1.south east)+(0,-0.25cm)$)},anchor=north east,
        width=.84\linewidth,
        height=5cm, 
        xlabel style = {align=center},
        xlabel={Signed RTD \\ \textcolor{red}{$\xleftarrow{\textrm{More Likely Google}}$}\hspace{1cm}\textcolor{blue}{$\xrightarrow{\textrm{More Likely Reddit}}$}},        
        yticklabels from table={\datatable}{x},
        ytick = {1,...,26},
        ylabel={},
        xmin=-0.5,xmax=0.5,
        enlarge y limits={abs=2pt},
        ymajorgrids,
        blue,
        yticklabel style={font=\scriptsize, black},
        every axis label/.append style ={black},
        every tick label/.append style={black},       
        axis line style={black},        
        boxplot/draw direction=x,
        boxplot/average={auto},
        boxplot/every average/.style={/tikz/mark=*,mark size=1},
    ]
\boxplotprepared{\datatable}{AskReddit}
\boxplotprepared{\datatable}{AutoNewspaper}
\boxplotprepared{\datatable}{dirtykikpals}
\boxplotprepared{\datatable}{dirtyr4r}
\boxplotprepared{\datatable}{uusbreakingnewsnet}
\boxplotprepared{\datatable}{HentaiAndRoleplayy}
\boxplotprepared{\datatable}{DirtySnapchat}
\boxplotprepared{\datatable}{teenagers}
\boxplotprepared{\datatable}{dirtypenpals}
\boxplotprepared{\datatable}{newsbotbot}
\boxplotprepared{\datatable}{relationshipadvice}
\boxplotprepared{\datatable}{SluttyConfessions}
\boxplotprepared{\datatable}{Roleplaybuddy2}
\boxplotprepared{\datatable}{rapefantasies}
\boxplotprepared{\datatable}{RoleplayHentai}
    \end{axis}    
\end{tikzpicture}
\end{minipage}    
     \vspace{-1cm}
\caption{Signed Rank Turbulence Divergence (RTD) for the most divergent subreddits and hashtags comparing results from SERP against Twitter/X (left) and Reddit (right). Subreddits and hashtags that are more likely to appear in SERP results are listed on top (\textcolor{red}{red}). Subreddits and hashtags that are more likely to appear in the nonsampled social media data are listed on the bottom (\textcolor{blue}{blue}).}
     \label{fig:commrtdreddit}
\end{minipage}
\end{figure*}
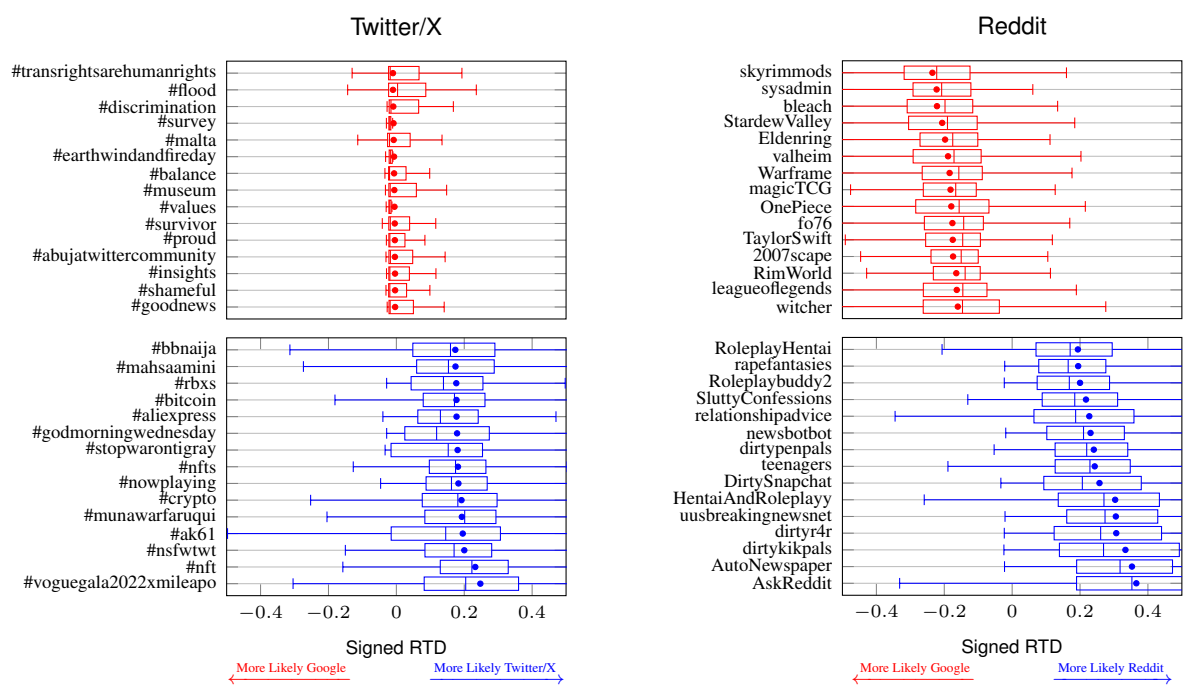

Figure~\ref{fig:commrtdreddit} shows the distributions of the 15 highest and lowest individual divergences (Eq. \ref{inverse_control}) and their mean (representing Eq. \ref{rtd}) for each subreddit and hashtag respectively. In other words, the subreddits and hashtags in red (\textit{i.e.}, top subplots) are more likely to be returned from Google's SERP than the nonsampled data and vice versa.
Because this analysis only looked at the extreme cases, and it is infeasible to visualize all the subreddits and hashtags in this manner, we further conducted a macro analysis, which provides more coverage into the subreddits and hashtags. See Section~\ref{sec:inclintion}.

\end{document}